\documentclass[10pt,twocolumn,letterpaper]{article}

\usepackage{cvpr}              

\usepackage{multirow}

\definecolor{cvprblue}{rgb}{0.21,0.49,0.74}
\usepackage[pagebackref,breaklinks,colorlinks,allcolors=cvprblue]{hyperref}
\usepackage{amsmath,amssymb}
\usepackage[accsupp]{axessibility}  

\def\paperID{21} 
\def\confName{CVPR}
\def\confYear{2026}

\title{The Surprising Effectiveness of Canonical Knowledge Distillation for Semantic Segmentation}

\author{
Muhammad Ali$^1$ \quad
Kevin Alexander Laube$^2$ \quad
Madan Ravi Ganesh$^2$ \quad
Lukas Schott$^3$ \\[2pt]
Niclas Popp$^{2,4}$ \quad
Thomas Brox$^1$ \\[4pt]
{\small $^1$University of Freiburg \quad
$^2$Bosch Center for Artificial Intelligence \quad
$^3$Aleph Alpha Research  \quad
$^4$ University of Tübingen}\\
{\tt\small alimu@informatik.uni-freiburg.de, 
Madan.RaviGanesh@us.bosch.com}
}

\newcommand{\increase}[1]{\textcolor{green!65!black}{\scriptsize\,($+$#1)}}

\begin{document}
\maketitle
\begin{abstract}

\noindent Recent knowledge distillation (KD) methods for semantic segmentation introduce increasingly complex hand-crafted objectives, yet are typically evaluated under fixed iteration schedules. These objectives substantially increase per-iteration cost, meaning equal iteration counts do not correspond to equal training budgets. It is therefore unclear whether reported gains reflect stronger distillation signals or simply greater compute.
We show that iteration-based comparisons are misleading:
when wall-clock compute is matched, \textit{canonical} logit- and feature-based KD outperform recent segmentation-specific methods.
Under extended training, feature-based distillation achieves state-of-the-art ResNet-18 performance on Cityscapes and ADE20K.
A PSPNet ResNet-18 student closely approaches its ResNet-101 teacher
despite using only one quarter of the parameters, reaching 99\% of
the teacher's mIoU on Cityscapes (79.0 vs.\ 79.8) and 92\% on ADE20K.
Our results challenge the prevailing assumption that KD for segmentation requires task-specific mechanisms and suggest that scaling, rather than complex hand-crafted objectives, should guide future method design.
\end{abstract}
\section{Introduction}
\label{sec:intro}

\noindent Semantic segmentation is central to applications such as autonomous
driving, robotics, and medical imaging, where dense scene
understanding must operate under tight memory and latency constraints.
Knowledge distillation (KD)~\cite{hinton-arxiv15a} offers an attractive
path to efficient models: a compact \textit{student} is trained to
reproduce the behaviour of a larger \textit{teacher}, without requiring
changes to architecture or precision.
The earliest and most widely adopted forms, logit-based
KD~\cite{hinton-arxiv15a} and feature-based KD
(FitNets~\cite{romero-iclr15a}), are simple, task-agnostic, and
computationally lightweight, requiring no auxiliary networks or
specialized objectives. We refer to these as \textit{canonical} methods.

\noindent Subsequent work in image classification KD proposed increasingly elaborate objectives:
relational methods~\cite{park-cvpr19a}, contrastive losses~\cite{tian-iclr20a},
and correlation-based objectives~\cite{peng2019correlation},
which improved performance but increased conceptual and computational complexity.
However, Hao \etal~\cite{hao-neurips23a} showed that when data and compute are scaled, Vanilla logit KD matches or outperforms recent KD methods on ImageNet~\cite{deng-cvpr09a}.

\noindent Semantic segmentation differs from image classification in that it predicts a label for every pixel, producing hundreds of thousands of structured outputs per image. This dense structure has motivated task-specific distillation objectives such as boundary-aware losses, memory-based relations, and masked feature generation~\cite{liu2024bpkd,yang-cvpr22a,yang-eccv22a}. Yet it remains unclear whether such complexity is fundamentally necessary, or whether canonical methods have simply been undertrained and under-evaluated.
Echoing Sutton’s “Bitter Lesson”~\cite{sutton2019bitter}, we ask whether segmentation distillation benefits more from scalable objectives than from increasingly specialized designs.

\noindent We address this question by examining segmentation KD from a compute-aware and long-horizon perspective. Current evaluation protocols compare methods at fixed iteration counts, e.g. 40k on Cityscapes~\cite{cordts-cvpr16a}, ignoring that more complex objectives often introduce additional computations or memory structures and can incur up to $3\times$ higher per-iteration cost. As a result, equal iteration counts do not correspond to equal training budgets. Moreover, these short schedules may not reflect realistic distillation practice. We therefore study how canonical and segmentation-specific methods scale as additional compute becomes available, extending training far beyond standard schedules. Our key findings are:
\begin{enumerate}
  \item \textbf{Iteration-count comparisons misrepresent method quality.} Hand-crafted segmentation KD objectives incur substantially higher per-iteration cost, making fixed-iteration benchmarks a poor proxy for training budget. When wall-clock compute is matched, canonical logit- and feature-based KD outperform more elaborate objectives.
  \item \textbf{Canonical KD scales strongly with additional compute.}
    Examining scaling behaviour across extended training budgets reveals that supervised learning saturates and degrades, while canonical KD converges later, reaches higher performance ceilings, and degrades more gradually. 
    Feature-based distillation produces state-of-the-art PSPNet ResNet-18 models on Cityscapes and ADE20K, reaching 99\% of its ResNet-101 teacher on Cityscapes and 92\% on ADE20K.
  \item \textbf{Feature and logit KD occupy complementary regimes.}
  Feature-based KD performs best in same-architecture, in-distribution settings, while logit-based KD scales more robustly to semi-supervised training with additional unlabeled images. This suggests that intermediate feature alignment is more sensitive to distribution shift, whereas logit supervision remains stable.
\end{enumerate}

\section{Background}
\label{sec:background}

\subsection{Logit-based distillation}

Logit-based KD~\cite{hinton-arxiv15a} trains the student to match the
teacher's softened output distribution.
Let $z^T_{i,j} \in \mathbb{R}^C$ and $z^S_{i,j} \in \mathbb{R}^C$
denote the teacher and student logits at spatial location $(i,j)$ for a
segmentation task with $C$ classes.
Applying a temperature-scaled softmax produces softened distributions
$p^T_{i,j,c}$ and $p^S_{i,j,c}$.
The distillation loss minimizes their KL divergence, averaged over all
spatial positions:
\begin{equation}
\mathcal{L}_{\text{KD}} =
\frac{T^2}{HW}
\sum_{i,j} \sum_{c=1}^C
p^T_{i,j,c} \log \frac{p^T_{i,j,c}}{p^S_{i,j,c}},
\end{equation}
where $T$ is the temperature.

\subsection{Feature-based distillation}

Feature-based KD (FitNets~\cite{romero-iclr15a}) aligns intermediate
representations of the teacher and student networks.
Let $F^T \in \mathbb{R}^{H \times W \times C_T}$ and
$F^S \in \mathbb{R}^{H \times W \times C_S}$ denote teacher and student
feature maps at a given layer.
A learnable $1 \times 1$ projection $r(\cdot)$ maps student features
to the teacher's channel dimension. The loss is:
\begin{equation}
\mathcal{L}_{\text{feat}} =
\frac{1}{HW}
\sum_{i,j}
\left\| r(F^S_{i,j}) - F^T_{i,j} \right\|_2^2.
\end{equation}
Feature matching can be applied at multiple encoder stages
When combined with logit-based KD, the overall objective becomes
$\mathcal{L}_{\text{total}} = \mathcal{L}_{\text{CE}} +
\alpha\,\mathcal{L}_{\text{KD}} + \beta\,\mathcal{L}_{\text{feat}}$.

\subsection{Hand-crafted distillation methods}

We compare against several segmentation-specific methods proposed in the literature.
CIRKD~\cite{yang-cvpr22a} maintains cross-image memory queues to capture
global semantic relations.
SKD~\cite{liu-cvpr19skd}, CWD~\cite{shu-iccv21a}, and
MGD~\cite{yang-eccv22a} distill structured pixel relations,
channel-normalised feature maps, and masked feature generation,
respectively.
BPKD~\cite{liu2024bpkd} applies separate losses to boundary and interior
regions.
LAD~\cite{liu-wacv24a} decomposes feature matching into magnitude and
angular components.

\section{Benchmarking Setup}
\label{sec:setup}

\subsection{Datasets and models}

We evaluate on two widely-used benchmarks.
\textbf{Cityscapes}~\cite{cordts-cvpr16a} contains 2,975 training and
500 validation images of urban driving scenes at $2048\times1024$
resolution, with 19 semantic classes.
\textbf{ADE20K}~\cite{zhou-cvpr17a} provides 20,210 training and 2,000
validation images from diverse indoor and outdoor scenes, 
annotated with 150 categories.

\noindent Following earlier studies, we use PSPNet~\cite{zhao-cvpr17a} with a ResNet-101
backbone~\cite{he-cvpr16a} as the teacher (79.8\% mIoU on Cityscapes,
44.4\% on ADE20K) and PSPNet with a ResNet-18 backbone as the student ($\sim$4$\times$ fewer parameters).
Both backbones are initialised from ImageNet pre-trained weights.
For cross-architecture experiments, we additionally use a Mask2Former~\cite{cheng-cvpr22a}
teacher with a Swin backbone~\cite{liu2021swin} (83.7\% mIoU on Cityscapes).

\subsection{Training protocols}

\textbf{Short horizon: }
We evaluate two standard short-horizon training protocols.
The 80k-iteration setting used in standard protocol comparisons follows the training procedures of LAD~\cite{liu-wacv24a} and BPKD~\cite{liu2024bpkd} (batch size 16, SGD,  poly LR with initial value 0.02, PSPNet-R101 teacher), which we replicate as closely as possible.
The 40k-iteration setting follows the official CIRKD codebase~\cite{yang-cvpr22a} (DeepLabV3-R101 teacher, initial LR 0.01). For canonical methods, we set $\alpha = 1$ and $\beta = 6$ in all experiments.

\noindent \textbf{Long horizon:}
Standard short-horizon recipes can overfit when training is extended~\cite{beyer-arxiv22a}. We therefore develop a revised protocol with
stronger regularization: we add photometric distortion to the
standard augmentation pipeline, replace the polynomial LR schedule with cosine annealing and warmup, and tune hyperparameters with a small grid search at each budget.
\footnote{Relative to the LAD/BPKD recipe, our protocol improves supervised learning by $+2.1$ mIoU, logit KD by $+1.2$, and feature KD by $+0.6$ on Cityscapes after 140k iterations.} 
\section{Results}
\label{sec:results}


\subsection{Short Horizon: Comparison Under Standard Protocols}
\label{sec:sota-short}

We first compare against recent SOTA segmentation KD methods under their published fixed-iteration training schedules (80k iterations on Cityscapes, 160k on ADE20K). Because baselines vary across implementations, we report improvements relative to each method's own baseline. Feature-based KD achieves the strongest improvement on both benchmarks (+4.4\% on Cityscapes, +5.4\% on ADE20K for PSPNet-R18), demonstrating that canonical methods are highly competitive even under standard short-horizon protocols.

\begin{table}[t]
\centering
\caption{Comparison with SOTA KD methods at 80k/160k iterations using a PSPNet-R18 student and a PSPNet-R101 teacher.
Parentheses show mIoU gain over each method's own baseline.
\textbf{Bold}: best gain; \underline{underline}: second best.
As the baselines vary across codebases, the results are not directly comparable. Results for SKD, CWD, MGD and LAD are from \cite{liu-wacv24a}, which reimplements methods without public code.
}
\label{tab:80k-results}
\small
\setlength{\tabcolsep}{3pt}
\begin{tabular}{l cc cc}
\toprule
\multirow{2}{*}{\textbf{Method}}
  & \multicolumn{2}{c}{\textbf{Cityscapes}}
  & \multicolumn{2}{c}{\textbf{ADE20K}} \\
\cmidrule(lr){2-3} \cmidrule(lr){4-5}
  & Base & Distilled & Base & Distilled \\
\midrule
T: PSP-R101 & 79.8 & -- & 44.4 & -- \\
\midrule
SKD~\cite{liu-cvpr19skd}   & 72.7 & 74.2\,{\increase{1.5}} & 35.0 & 35.3\,{\increase{0.3}} \\
CWD~\cite{shu-iccv21a}     & 72.7 & 75.9\,{\increase{3.2}} & 35.0 & 36.8\,{\increase{1.8}} \\
MGD~\cite{yang-eccv22a}     & 72.7 & 75.9\,{\increase{3.2}} & 35.0 & 36.8\,{\increase{1.8}} \\
LAD~\cite{liu-wacv24a}      & 72.7 & 76.9\,{\increase{\underline{4.2}}} & 35.0 & 39.6\,{\increase{4.6}} \\
BPKD~\cite{liu2024bpkd}    & 74.2 & 77.6\,{\increase{3.4}} & 33.3 & 38.5\,{\increase{\underline{5.2}}} \\
\midrule
Logit KD (Canonical)        & 73.4 & 74.1\,{\increase{0.7}} & 33.5 & 33.7\,{\increase{0.3}} \\
Feature KD (Canonical)         & 73.4 & 77.8\,{\increase{\textbf{4.4}}} & 33.5 & 38.8\,{\increase{\textbf{5.4}}} \\
\bottomrule
\end{tabular}
\end{table}

\subsection{Short Horizon: Compute-aware Comparison}
\label{sec:compute-aware}


We now revisit short-horizon evaluation under matched compute budgets. We compare canonical methods against CIRKD~\cite{yang-cvpr22a}, a strong segmentation-specific KD method, using its official codebase and 40k-iteration training protocol on Cityscapes. While CIRKD attains 73.5\% mIoU after 30.2 GPU hours, feature-based KD completes 40k iterations in only 9.2 GPU hours with comparable performance (73.7\%).

\noindent \Cref{fig:compute-aware} contrasts iteration-matched and compute-matched comparisons. When training budgets are matched in wall-clock time, canonical methods outperform CIRKD: logit KD by 2.9 mIoU and feature KD by 1.9 mIoU.


\begin{figure}[t]
    \centering
    \includegraphics[width=\linewidth]{figures/40k_compute_vs_iter.pdf}
    \caption{Iteration-matched vs.\ compute-matched comparison. CIRKD appears competitive under iteration-matched evaluation but is outperformed when compute is matched.}
    \label{fig:compute-aware}
\end{figure}

\subsection{Scaling Behaviour Under Extended Compute}
\label{sec:long-horizon}


\noindent We extend training up to 300 GPU hours to examine how different distillation strategies scale as additional compute is made available. For CIRKD, we use its official DeepLabV3-R101 teacher, which achieves slightly higher mIoU than PSPNet-R101.

\begin{figure}[t]
    \centering
    \includegraphics[width=\linewidth]{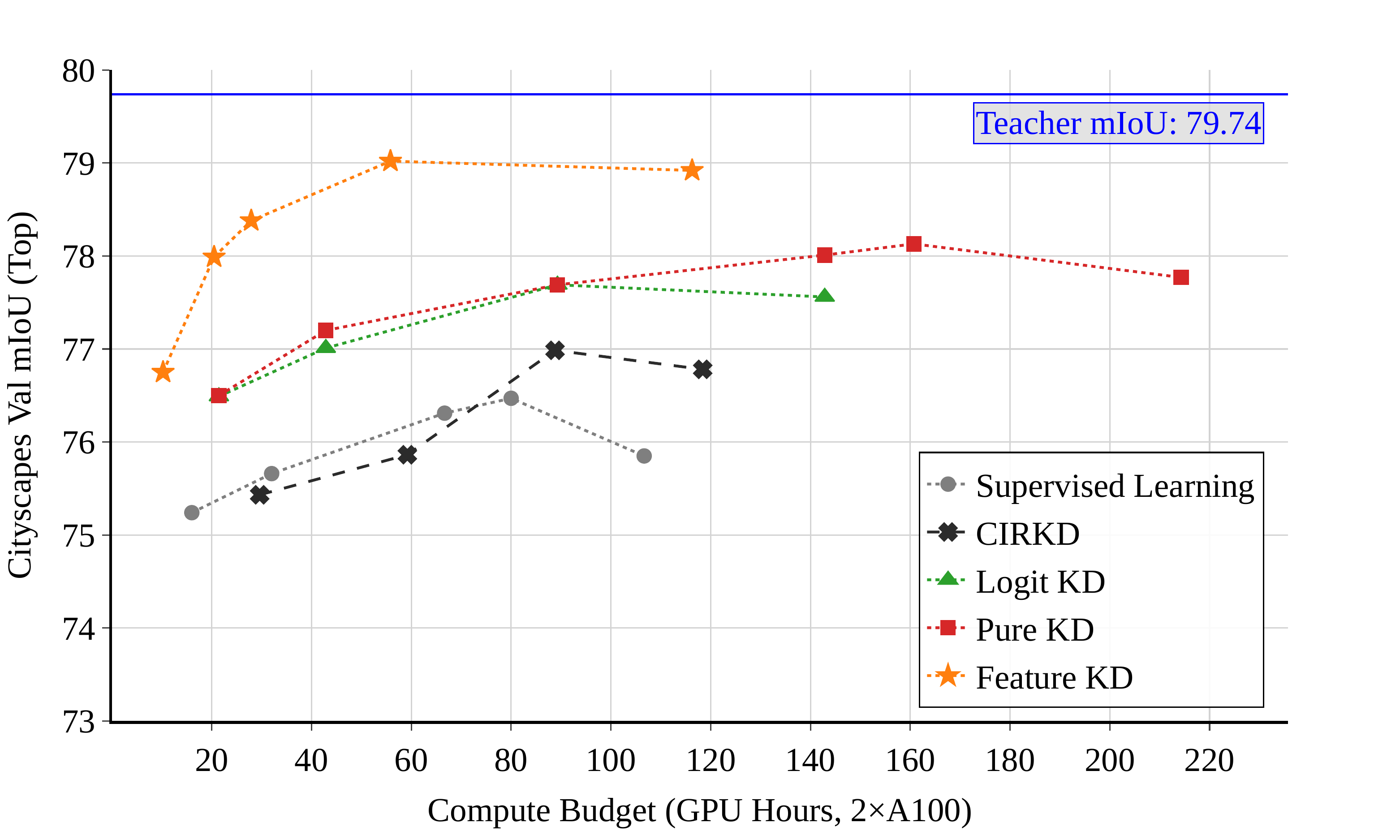}
    \caption{mIoU vs training budget on Cityscapes under long-horizon training. Each point shows the best validation performance achieved at a given GPU-hour budget across hyperparameter configurations.}
    \label{fig:long-horizon}
\end{figure}

\noindent \Cref{fig:long-horizon} plots the mIoU as a function of training budget. Each point represents the best validation performance achieved at that budget across hyperparameter settings. \Cref{tab:long-horizon-results} reports the peak performance achieved by each method.


\begin{table}[t]
    \centering
    \caption{Best validation mIoU under long-horizon training.}
    \label{tab:long-horizon-results}
    \small
    \setlength{\tabcolsep}{3pt}
    \begin{tabular}{l ccc ccc}
    \toprule
    \multirow{2}{*}{\textbf{Method}}
      & \multicolumn{3}{c}{\textbf{Cityscapes}}
      & \multicolumn{3}{c}{\textbf{ADE20K}} \\
    \cmidrule(lr){2-4} \cmidrule(lr){5-7}
      & mIoU & Epochs & GPUh
      & mIoU & Epochs & GPUh \\
    \midrule
    T: PSP-R101 & 79.8 & -- & -- & 44.4 & -- & -- \\
    \midrule
    Supervised  & 76.3 & 2500 & 67  & 37.7 & 1500 & 250 \\
    + CIRKD     & 77.0 & 645 & 44  & 40.7 & 570 & 260 \\
    + Logit KD  & 77.7 & 2500 & 89  & 39.4 & 1200 & 240 \\
    + Pure KD   & 78.1 & 4500 & 160 & 38.8 & 1200 & 240 \\
    + Feature KD & \textbf{79.0} & 1200 & 55 & \textbf{41.1} & 600 & 157 \\
    \bottomrule
    \end{tabular}
\end{table}

\noindent Supervised learning saturates and degrades under extended training, whereas KD methods peak later, achieve higher performance, and degrade more gradually. Feature-based KD achieves the strongest results on both benchmarks:
79.0\% mIoU on Cityscapes (99\% of teacher) in only 55 GPU hours, and
41.1\% on ADE20K (92\% of teacher).
Logit-based KD and Pure KD (logit KD with only KD loss, no ground-truth supervision) also scale favourably, reaching 77.7\% and
78.1\% respectively on Cityscapes, showing that canonical
methods continue improving beyond standard schedules.

\subsection{Further Analysis}
\label{sec:further}

\textbf{Semi-supervised KD:}
Since teacher predictions provide supervision without ground-truth labels, KD naturally extends to unlabeled data. Using the Cityscapes Train Extra split (20{,}000 additional images, labels ignored), Pure KD scales strongly, reaching 79.0\% mIoU after 320 GPU hours—matching the best fully-supervised feature KD result (\Cref{tab:long-horizon-results}) and 99\% of teacher mIoU. In contrast, feature-based KD does not benefit from the additional data, suggesting that intermediate feature alignment is more sensitive to distribution shift. These results indicate that logit-based KD is better suited to semi-supervised or data-heterogeneous settings.

\begin{figure}[t]
    \centering
    \includegraphics[width=\linewidth]{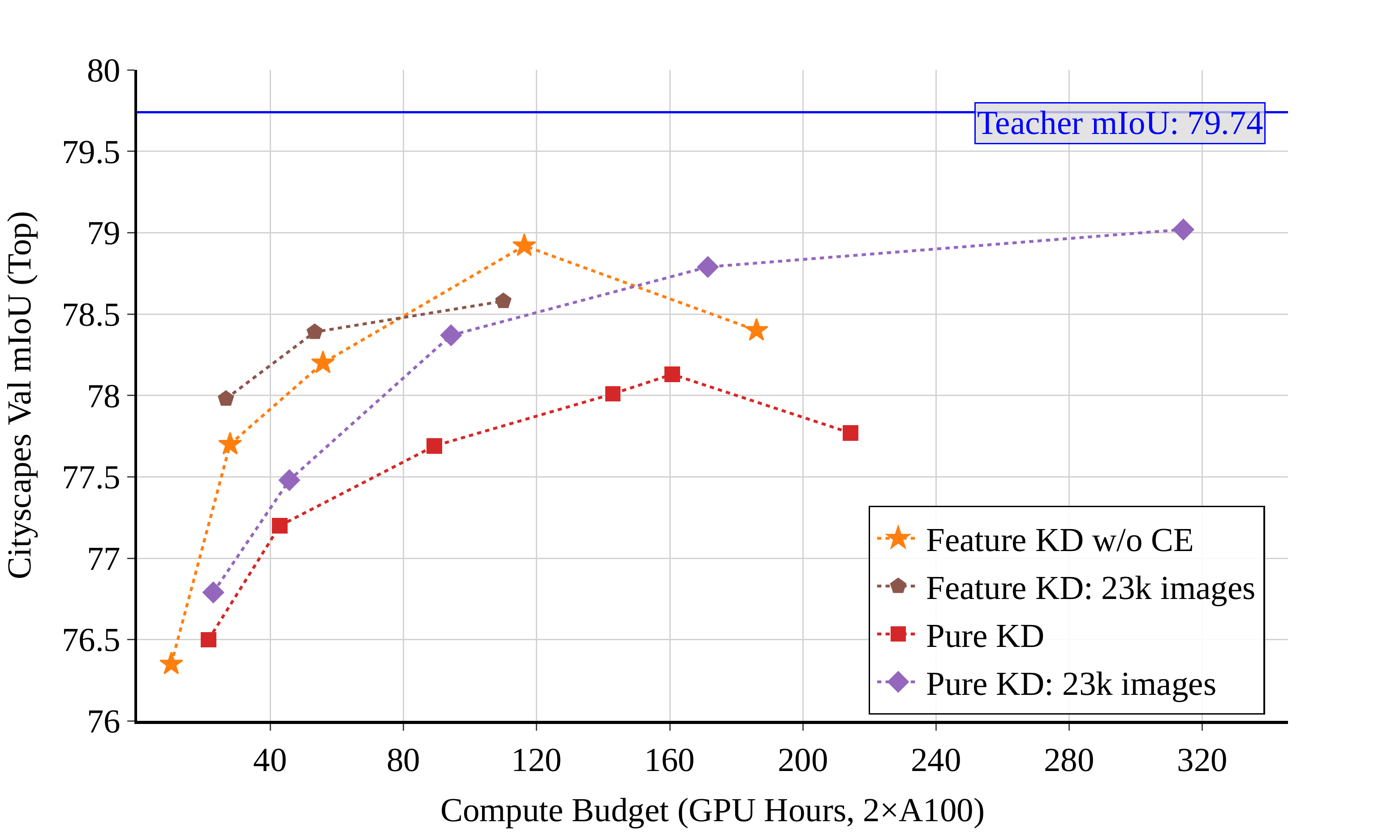}
    \caption{Semi-supervised KD on Cityscapes with 20k additional unlabelled images. Pure KD (logit KD without CE) scales to 79.0\% mIoU, matching
    feature KDs' best fully-supervised result, while feature KD does not improve.}
    \label{fig:semi-supervised}
\end{figure}

\noindent \textbf{Teacher architecture:}
On Cityscapes, a Mask2Former teacher (83.7\% mIoU) yields only a
marginal gain over PSPNet-R101 (77.8\% vs.\ 77.7\% student mIoU) at $3\times$ the compute (296 vs.\ 89 GPU hours).
On ADE20K, it produces weaker students than the architecturally aligned PSPNet-R101, suggesting that architectural compatibility matters more than raw
teacher accuracy for dense prediction.

\section{Discussion and Conclusion}
\label{sec:discussion}

\textbf{Why do canonical methods work so well?}
Semantic segmentation provides extremely dense supervision. A single Cityscapes image produces over half a million predictions. This may already encode much of the structural and contextual information that hand-crafted objectives attempt to impose explicitly, reducing the marginal benefit of auxiliary
relational or cross-image losses.
A second factor is the interaction between distillation and extended
training: while supervised learning saturates and degrades, KD continues
to improve, suggesting that teacher supervision provides a harder
objective that enables longer effective
optimization~\cite{beyer-arxiv22a}.
Viewed through the lens of Sutton's Bitter
Lesson~\cite{sutton2019bitter}, canonical objectives benefit
from additional compute, while hand-crafted methods encode task-specific
assumptions whose additional costs limits their scalability.

\noindent \textbf{Segmentation vs.\ classification KD:}
Our findings reflect recent results in classification, where scaled vanilla logit KD matches hand-crafted alternatives~\cite{hao-neurips23a}.
However, we observe key differences in our experiments.
The sensitivity to the softmax temperature parameter is markedly lower: we find $T\in[1,4]$ optimal,
whereas classification KD typically uses $T\in[4,20]$.
The dense supervision already provides rich gradients, and excessive
softening risks blurring spatial precision.
Feature-based KD also plays a larger role than in classification,
consistently outperforming logit KD in same-architecture settings,
suggesting that intermediate representations encode spatial cues that are
partially lost when compressed into final logits.

\noindent \textbf{Practical recommendations:} For same-architecture student--teacher pairs, feature-based KD offers the
best trade-off between performance and complexity, achieving near-teacher
accuracy with minimal tuning.
When architectures differ or additional unlabelled data is available, logit-based KD
is more robust, avoiding reliance on potentially misaligned intermediate
representations.
Extending training well beyond standard schedules yields
consistent gains; however, the benefit of KD must be weighed against the cost of training or acquiring a suitable teacher.

\noindent \textbf{Implications for future research:}
Our results suggest that new distillation objectives should be compared
against well-tuned canonical baselines under matched compute budgets,
not fixed iteration counts.
Future work should report scaling curves rather than single
operating points, as these reveal whether a method continues to improve,
plateaus, or degrades under extended training.
When performance differences are small, methods that are simpler to
implement and tune likely have greater practical
impact~\cite{hao-neurips23a}.

\noindent \textbf{Limitations:}
As our study focuses on CNN-based students with ResNet backbones, it remains unclear whether these findings extend to larger capacity gaps or diverse architectures.
Additionally, several recent methods could not be reimplemented under identical long-horizon conditions due to limited code availability;
a fully unified comparison would strengthen our conclusions.

\noindent \textbf{Conclusion:}
Canonical KD methods achieve state-of-the-art segmentation performance under realistic compute budgets and long training horizons.
A PSPNet ResNet-18 student reaches 99\% of its ResNet-101 teacher on Cityscapes, and logit-based KD matches this result in a semi-supervised setting without any ground-truth labels.
These findings challenge the assumption that segmentation requires task-specific distillation complexity, and suggest that scalability and simplicity should guide future method design.
\clearpage
{
    \small
    \bibliographystyle{ieeenat_fullname}
    \bibliography{main}
}


\end{document}